\documentclass[11pt]{article}
\usepackage[preprint]{acl}
\usepackage{times}
\usepackage{latexsym}
\usepackage[T1]{fontenc}
\usepackage{iftex}
\ifPDFTeX
  \usepackage[utf8]{inputenc}
\fi
\ifXeTeX\fi
\ifLuaTeX\pdfvariable minorversion=7\fi
\usepackage{microtype}
\usepackage{inconsolata}
\usepackage{graphicx}
\usepackage{amsmath}
\usepackage{booktabs}
\usepackage{placeins}

\title{ACVA-Bench: Beyond Prompt Following in Abstract Concept Visualization Ability for Text-to-Image Models}
\author{
\textbf{Mengyan Wang$^{1,2}$}\thanks{These authors contributed equally to this work.}\quad
\textbf{Jiyuan An$^{1,2}$}\footnotemark[1]\quad
\textbf{Yining Liu$^{1,2}$} \\
\textbf{Yaowen Zhang$^{1,2}$}\quad
\textbf{Liner Yang$^{1,2}$}\quad
\textbf{Erhong Yang$^{1,2}$} \\[0.5em]
\normalfont\small $^{1}$National Language Resources Monitoring and Research Center for Print Media, \\
\normalfont\small Beijing Language and Culture University, China \\
\normalfont\small $^{2}$School of Information Science, Beijing Language and Culture University, China
}

\begin{document}
\maketitle
\begin{abstract}
Text-to-image (T2I) models have advanced in image fidelity and instruction following, but their ability to visualize abstract concepts without concrete visual specifications remains unclear. We formalize Abstract Concept Visualization Ability (ACVA): translating concepts without fixed visual referents into accurate, diverse, and distinguishable images. We introduce ACVA-Bench, a category-balanced benchmark of 150 concepts grounded in psycholinguistic norms for Chinese abstract words. A low-intervention protocol supplies only the target concept, without specifying objects, scenes, or metaphors. The benchmark evaluates semantic adequacy, representational diversity, and conceptual distinctiveness. To accommodate multiple valid depictions, we construct dictionary-grounded visual-semantic rubrics that decompose concept meaning into independently judgeable questions and validate automatic scores against human judgments. Evaluating six T2I models on 2,700 images reveals substantial differences across models and cognitive categories. Semantically strong models often rely on a narrow set of visual prototypes, while surface variation can conceal unchanged metaphors. Closely related concepts with narrow semantic boundaries remain difficult to distinguish. ACVA-Bench provides an interpretable framework for studying the visual grounding of abstract meaning beyond explicit prompt following.

\end{abstract}

\section{Introduction}
\label{sec:intro}
\begin{figure*}[t]
\centering
\includegraphics[width=\linewidth]{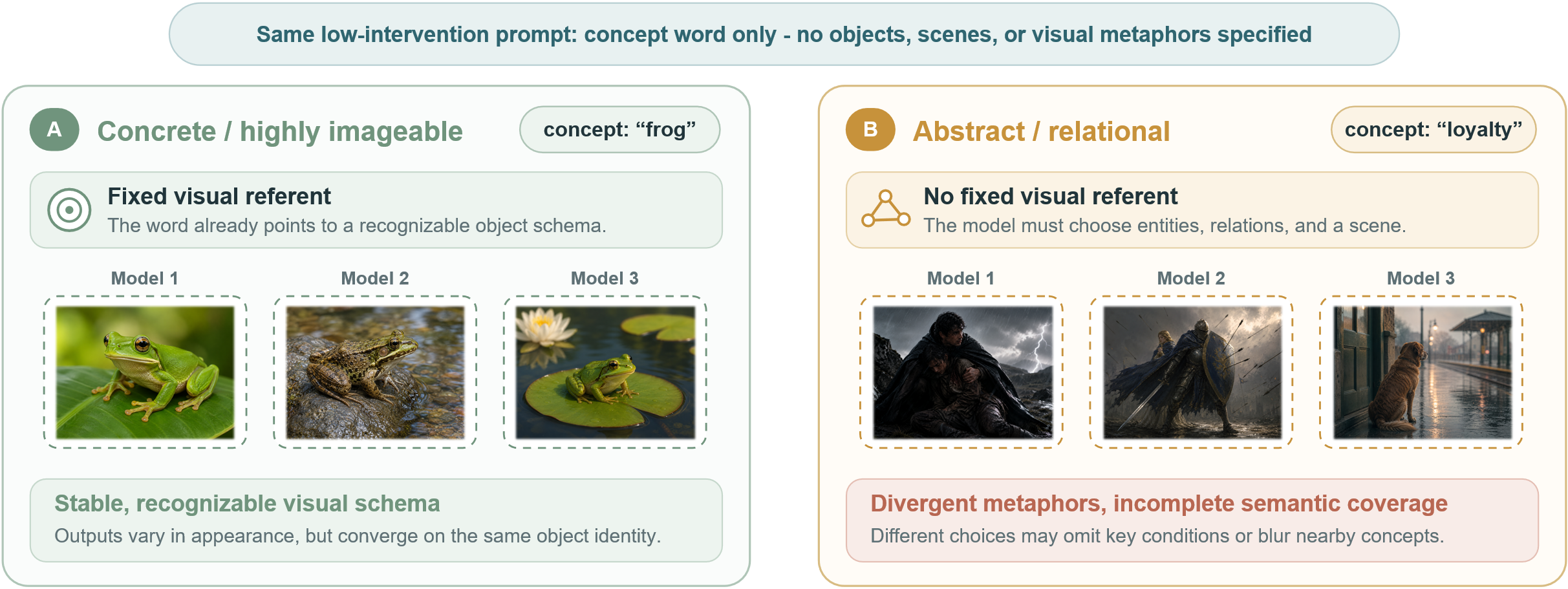}
\caption{\textbf{Motivation.} Concrete concepts have fixed visual referents, whereas abstract concepts require models to autonomously select visual entities, relations, and metaphors, often resulting in divergent and incomplete representations.}
\label{fig:motivation}
\end{figure*}

Text-to-image (T2I) models have advanced in image quality, scene generation, and instruction following \citep{rombach2022,saharia2022,esser2024}. Recent evaluations probe implicit meaning and reasoning, but still provide elaborate prompts containing task constraints or reasoning cues \citep{sun2026}. This leaves a distinct question: can a model autonomously choose the entities, relations, and scenes needed to express a bare abstract concept? Unlike concrete concepts such as \emph{frog}, abstract concepts such as \emph{loyalty} lack a fixed visual referent. Models must select a visual realization, and their depictions can diverge or omit essential meaning (Figure~\ref{fig:motivation}).

We define Abstract Concept Visualization Ability (ACVA) as the ability to organize visual entities, relations, scenes, and schemas into recognizable, diverse representations that distinguish related concepts, without an explicit visual implementation plan. This connects visual generation with a linguistic problem: grounding abstract meaning while preserving its semantic conditions.

Fine-grained evaluators such as TIFA and EvalMuse \citep{hu2023tifa,han2026evalmuse} assess visual facts specified in advance. Abstract meaning admits multiple valid situations and metaphors, so it requires a reference that constrains semantics without prescribing objects. Moreover, abstract concepts draw differently on sensorimotor, emotional, and social experience \citep{harpaintner2018,jiang2026}; evaluation should cover this heterogeneity.

We introduce ACVA-Bench, a category-balanced benchmark grounded in psycholinguistic norms for Chinese abstract concepts. A low-intervention protocol specifies only the target word, without objects, scenes, or metaphors. Evaluation measures semantic adequacy, representational diversity, and conceptual distinctiveness. Dictionary definitions, componential analysis, and frame semantics guide independently judgeable visual-semantic questions; automatic scores are compared with human judgments.

Our contributions are threefold: (1) formalizing ACVA as autonomous visual grounding of abstract semantics; (2) constructing a cognitively balanced benchmark with three complementary evaluation axes; and (3) developing an interpretable, feature-level scoring method and applying it to six image-generation models with human validation.

\section{Related Work}
\label{sec:related}
\subsection{Multidimensional Cognitive Representations of Abstract Concepts}
Early work often described concepts with a strict concrete-abstract dichotomy, but growing evidence suggests abstract concepts are not a homogeneous category. The multiple-representation view holds that conceptual knowledge is jointly supported by partially overlapping systems -- sensorimotor, linguistic, emotional, interoceptive, social-interactive, and introspective -- to different degrees across concepts \citep{barsalou2005situating,troche2014,conca2021}. Feature-rating and clustering studies provide empirical grounding for this heterogeneity \citep{harpaintner2018,villani2019}, and recent cross-linguistic work extends such norming to 503 Chinese abstract words, revealing core cognitive dimensions and categories \citep{jiang2026}. This heterogeneity has been validated mainly in human concept processing \citep{borghi2018,banks2023}; work on generative models still largely treats abstract concepts as undifferentiated.

\subsection{Visualizing Abstract Concepts with Generative Models}
Existing work relates abstract concepts to images mainly via generation assistance and visual-representation analysis. TIAC and POAC use a language model to map abstract concepts onto concrete objects and scenes, constructing a prompt a T2I model can execute \citep{liao2024,fan2024}; \citet{cerini2024} study indirect visual grounding through human- and LLM-generated situations with image-text retrieval. \citet{tater2024,tater2025} examine visual variability of abstract versus concrete concepts and build AbsVis to compare human/VLM depiction preferences; \citet{naber2025} find cross-modal alignment depends on concept concreteness, and \citet{gunther2022} project text-based representations into visual space for unseen abstract words. These studies show abstract concepts can obtain indirect visual grounding via a language model's semantic parsing, but whether a T2I model can organize visual entities and relations by itself remains unexamined -- the gap our work addresses.

\subsection{Text-to-Image Evaluation}
T2I evaluation has gradually shifted from holistic image quality toward fine-grained analysis of specific capabilities. FID and CLIPScore measure generation quality and image-text compatibility via feature distributions and vision-language representations \citep{heusel2017,hessel2021}; DrawBench evaluates quality and alignment via challenging prompts and human comparison, while HEIM assesses performance through a multidimensional framework \citep{saharia2022,lee2023}. To localize compositional errors, GenEval evaluates object co-occurrence, count, color, and spatial position, and T2I-CompBench extends coverage to attribute binding and spatial, non-spatial, and complex relations \citep{ghosh2023,huang2023compbench}. TIFA and DSG decompose prompts into verifiable visual propositions checked via visual question answering, and EvalMuse-40K trains an automatic evaluator on fine-grained human annotation \citep{hu2023tifa,cho2024,han2026evalmuse}. T2I-ReasonBench evaluates whether models can interpret idioms, design requirements, entity relations, and scientific questions using LLM-generated question-criterion pairs \citep{sun2026}. These methods improve the granularity of T2I evaluation but measure fidelity to predetermined visual facts and rarely target abstract concepts -- the gap our VQA-style method addresses without reducing a concept to a single symbol.

\section{ACVA-Bench}
\label{sec:benchmark}
\begin{figure*}[t]
\centering
\includegraphics[width=\linewidth]{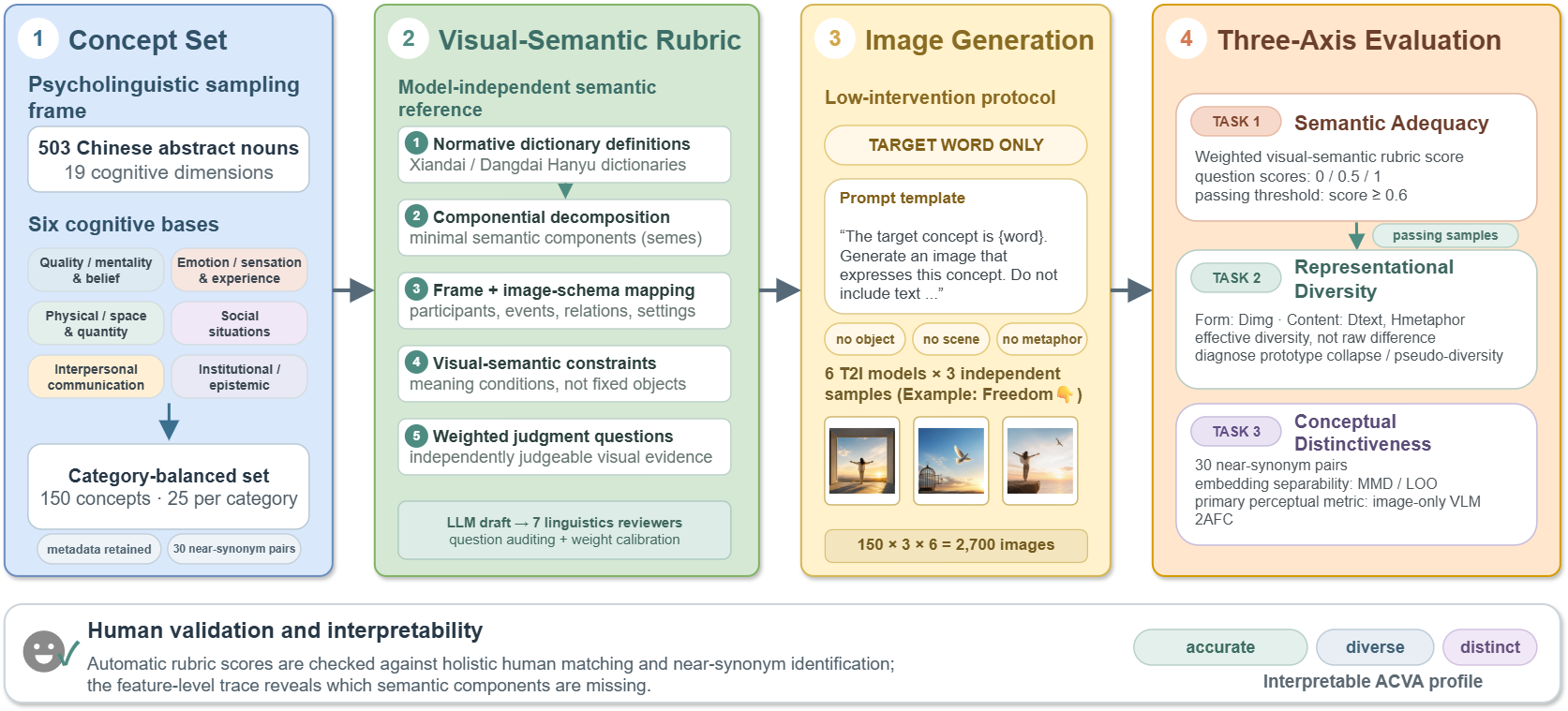}
\caption{Overview of ACVA-Bench: concept sampling, rubric construction, low-intervention generation, and three evaluation axes, with human quality control.}
\label{fig:framework}
\end{figure*}

ACVA-Bench comprises concept sampling, rubric construction, low-intervention image generation, and three-axis evaluation with human validation (Figure~\ref{fig:framework}). Psycholinguistic norms provide a category-balanced sampling frame; dictionary definitions guide model-independent semantic rubrics. Each model then generates multiple images from a target concept alone. Section~\ref{sec:metrics} defines the evaluation axes.

The source norms cover multiple cognitive dimensions. Although the benchmark uses Chinese concepts, the construction procedure could be applied to other languages with appropriate resources and validation.

\subsection{Abstract Concept Set}
\label{sec:concepts}
Candidate concepts are drawn from psycholinguistic norms for Chinese abstract concepts built by \citet{jiang2026} along 19 cognitive dimensions, covering 503 Chinese abstract nouns. Combining the original clustering results with our evaluation goals, we cover six categories -- quality, mentality, and belief; emotion, sensation, and experience; physical, spatiotemporal, and quantitative; social-situation; interpersonal-communication; and terminological, institutional, and epistemic concepts -- and stratify-sample 25 concepts from each, yielding a category-balanced list of 150 abstract concepts. We also retain each concept's original psycholinguistic ratings, such as imageability, as an external reference for later analysis. Within each stratum we construct five near-synonym pairs, giving 30 pairs in total. ``Near-synonym'' here does not require strict synonymy, only that the two words belong to the same or an adjacent semantic field and are strongly related overall. Pairs were verified against the subclass groupings in \emph{Tongyici Cilin} (an authoritative thesaurus classifying Chinese vocabulary by semantic category) or Chinese near-synonym discrimination references, to ensure their relatedness is well grounded.

\subsection{Visual-Semantic Rubric Resource}
\label{sec:rubric}
\begin{figure}[t]
\centering
\includegraphics[width=\linewidth]{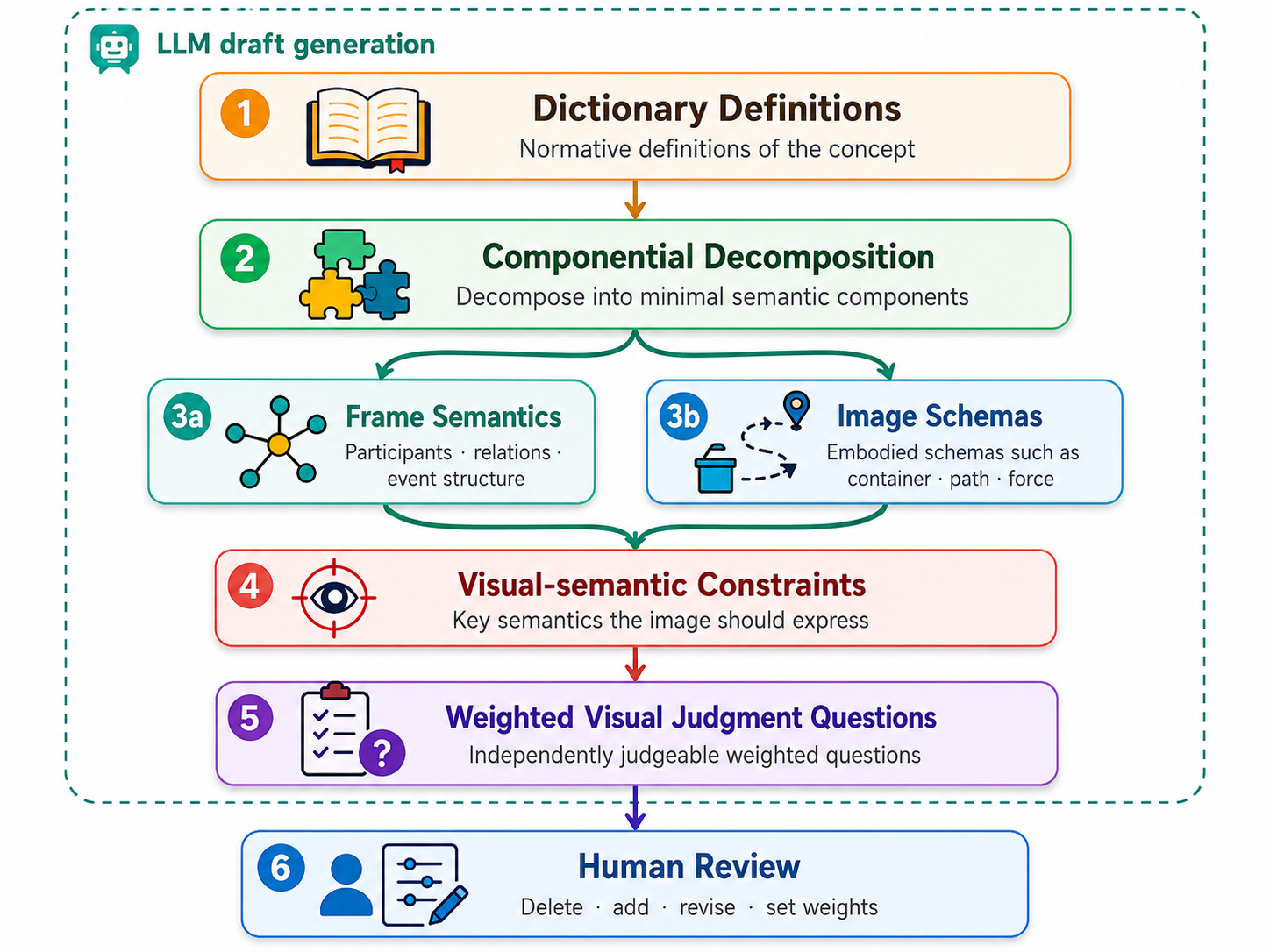}
\caption{Rubric construction pipeline from dictionary definitions to human-reviewed visual judgment questions. Constraints encode necessary semantic conditions rather than fixed objects or symbols; weights are human-calibrated.}
\label{fig:rubric}
\end{figure}

Because abstract concepts lack a unique reference image, we propose a fine-grained visual-semantic framework constrained by concept semantic structure, establishing a model-independent semantic reference for each concept. Figure~\ref{fig:rubric} shows the construction pipeline: dictionary definition $\rightarrow$ componential decomposition $\rightarrow$ frame/image-schema mapping $\rightarrow$ visual-semantic constraints $\rightarrow$ weighted visual judgment questions. The pipeline first generates draft data with an LLM, followed by human review and weight calibration, supported by prior work showing LLMs can generate human-like semantic features \citep{hansen2022semantic} and decompose text into verifiable semantic units \citep{hu2023tifa,cho2024}.

We extract definitions from the \emph{Xiandai Hanyu Cidian} (Contemporary Chinese Dictionary) and \emph{Dangdai Hanyu Xuexi Cidian} (Contemporary Chinese Learner's Dictionary). Componential (semic) analysis decomposes each definition into minimal semantic components. Frame semantics organizes their structural relations---who does what to whom, under what circumstances---as participants, events, relations, and settings. These map to image entities, relations, and scene layout. Image schemas such as container, path, force, up--down, and center--periphery further connect abstract relations to perceptible spatial structure. The resulting constraints specify semantic conditions rather than fixed objects or symbols; each becomes an independently judgeable visual question.

\subsection{Rubric Question Auditing}
\label{sec:auditing}
The generated rubric was reviewed by seven linguistics graduate students, who deleted, added, modified, and reweighted semantic features. We removed features unrelated to the target sense, subjective psychological inferences that cannot be judged from observable visual evidence, and semantically redundant features; we also revised question wording and finalized the weight of each feature.

For example, the rubric for \emph{loyalty} asks whether an image depicts an ongoing relationship and sustained support or allegiance, rather than mere proximity. A depicted act of protection or support can provide evidence without requiring one specific object, gesture, or setting. Conversely, an attractive scene of two people standing together need not express the relation's defining properties. This distinction makes the rubric reusable across different visual metaphors: it fixes the semantic questions while leaving the model to choose how to answer them visually. Appendix~\ref{app:resources} gives the worked example and category definitions.

\section{Evaluation Methodology: Axes and Metrics}
\label{sec:metrics}
Holistic image--text similarity cannot identify which semantic components an image expresses, whether repeated generations reuse one metaphor, or whether nearby concepts remain distinguishable. We assess these questions along three complementary axes.

\subsection{Semantic Adequacy}
\label{sec:adequacy}
Semantic adequacy measures whether an image generated from a target word expresses the concept's key semantic components. For each rubric question (Section~\ref{sec:rubric}), the evaluator assigns 0 (not expressed), 0.5 (partially expressed), or 1 (fully expressed). The image score is the weighted sum divided by the sum of human-calibrated weights (Section~\ref{sec:auditing}), giving a value in $[0,1]$. Images scoring at least 0.6 are \emph{passing samples}.

Formally, let $Q_c$ be the rubric for concept $c$, $w_{cq}$ the calibrated weight of question $q$, and $r_q(I,c)$ its rating for image $I$. Then
\begin{equation}
S(I,c)=\frac{\sum_{q\in Q_c}w_{cq}\,r_q(I,c)}{\sum_{q\in Q_c}w_{cq}}.
\label{eq:adequacy}
\end{equation}
Normalization accommodates rubrics with different numbers of questions. The question-level ratings retain a diagnostic trace: two images with similar aggregate scores can miss different semantic components, which can be inspected directly.

Unlike multiplying CLIP similarity by an aesthetic score \citep{liao2024}, this feature-level measure traces missing semantic components. It separates visual attractiveness from expression of concept meaning.

\subsection{Representational Diversity}
\label{sec:diversity}
Effective diversity concerns variation among valid depictions. We compute it for model$\times$concept groups with at least two passing images, and also report unfiltered results to expose the effect of selection.

Formal diversity $D_{img}$ is the mean pairwise cosine distance between image embeddings within a group. For content diversity, a vision-language model extracts each image's core visual-metaphor description. We embed these descriptions and compute their mean pairwise distance $D_{text}$. Clustering metaphor embeddings by a similarity threshold yields metaphor entropy $H_{metaphor}$, the number of clusters, and the share of the largest cluster. Distances are normalized by the standard deviation of within-group pairwise distances across the full corpus.

The corpus medians of normalized $D_{img}$ and $D_{text}$ define four quadrants. High image distance with low metaphor diversity indicates \emph{pseudo-diversity}: surface appearance varies while the metaphorical core remains unchanged. Together with the low--low quadrant, it diagnoses \emph{prototype collapse}, or reliance on a narrow set of depictions, which image distance alone can miss.

\subsection{Conceptual Distinctiveness}
\label{sec:distinctiveness}
Smaller semantic distances between concepts make their visual differentiation more demanding \citep{crutch2013}. We reuse independently generated images for the 30 near-synonym pairs (Section~\ref{sec:concepts}), such as \emph{belief}/\emph{faith}. No definitions, paired prompts, or additional images are supplied: the task tests distinctions activated by each word alone.

On the generative side, we measure image-distribution separability using RBF-kernel maximum mean discrepancy (MMD), centroid cosine distance, and leave-one-out nearest-centroid classification. We also classify embedded image captions using leave-one-out evaluation in text space. The primary, perceptual-side metric is a two-alternative forced choice: a judge unaware of the image's generating word selects between candidate concepts $\{A,B\}$ using image content alone. Accuracy near 0.5 indicates chance-level discrimination. Embedding separability and perceptual readability need not agree, so both are reported separately.

\section{Experiments}
\label{sec:experiments}
\noindent\textbf{Models and generation.} We evaluate six strong T2I models spanning non-open-weight systems---GPT Image 2 \citep{openai2026gptimage2}, Gemini 3.1 Flash Image \citep{googledeepmind2026gemini31flashimage}, and Qwen-Image-2.0 \citep{zhao2026qwenimage20}---and open-weight systems---GLM-Image \citep{zai2026glmimage}, Qwen-Image \citep{wu2025qwenimage}, and NextStep-1.1 \citep{stepfun2026nextstep11}. Each receives the same prompt: \textit{The target concept is ``\{word\}.'' Please generate an image that expresses this concept. Do not include any text in the image.} Three independent generations per concept yield 450 images per model and 2,700 overall.

\noindent\textbf{Evaluation.} Qwen3.6-27B scores each visual-semantic question using the three-point criterion in Section~\ref{sec:adequacy}; calibrated weights produce the image-level score. Qwen3-VL-Embedding-8B and Qwen3-Embedding-8B encode images and metaphor descriptions, respectively; GLM-5V-Turbo performs perceptual two-alternative discrimination. The benchmark package includes 700+ human-reviewed weighted questions, standardized prompts, model versions and generation parameters, images, metadata, and scores. Appendices~\ref{app:implementation} and~\ref{app:human-validation} give the full protocols.

\subsection{Semantic Adequacy: Model and Category Differences}

\begin{table}[t]
\centering
\small
\begin{tabular}{lccc}
\hline
Model & Mean & SD & Pass (\%) \\
\hline
GPT-Image-2 & \textbf{0.874} & 0.191 & \textbf{90.4\%} \\
Gemini-3.1-Flash & 0.781 & 0.288 & 79.6\% \\
Qwen-Image-2.0 & 0.556 & 0.378 & 53.6\% \\
GLM-Image & 0.531 & 0.373 & 48.7\% \\
Qwen-Image & 0.170 & 0.301 & 12.7\% \\
NextStep-1.1 & 0.155 & 0.278 & 10.2\% \\
\hline
Overall & 0.511 & 0.412 & 49.2\% \\
\hline
\end{tabular}
\caption{Semantic adequacy by model (weighted visual-semantic rubric score; $n=450$ per model). Pass denotes the percentage scoring at least 0.6.}
\label{tab:adequacy}
\end{table}

\begin{figure}[t]
\centering
\includegraphics[width=\linewidth]{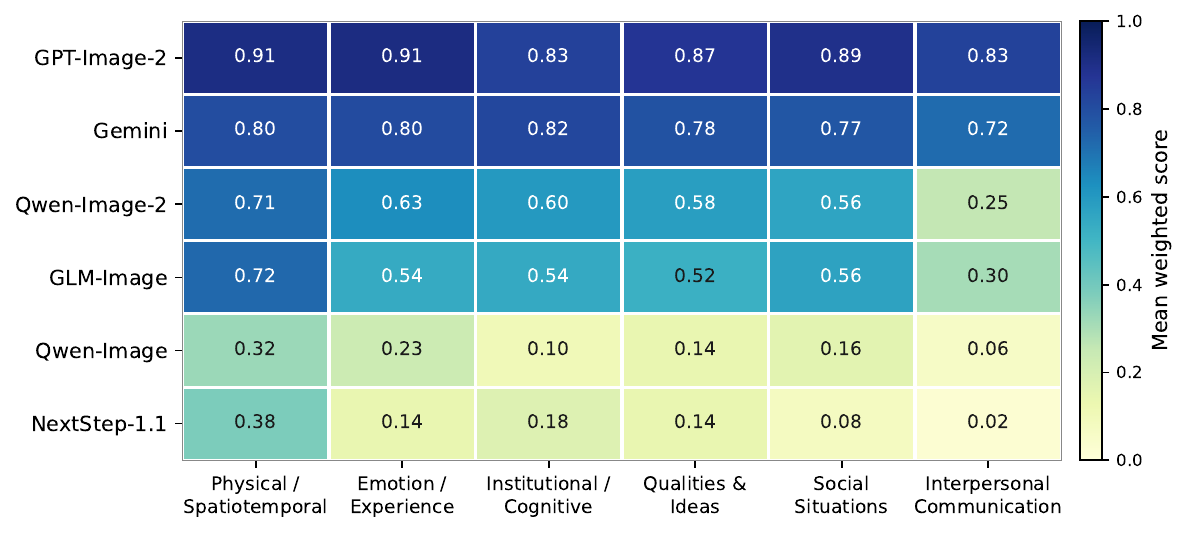}
\caption{Adequacy by model and cognitive category (75 images per cell). Interpersonal communication has the lowest mean across models.}
\label{fig:category}
\end{figure}

\noindent\textbf{Large capability gaps.} Table~\ref{tab:adequacy} shows that generating attractive images does not ensure adequate expression of abstract concepts. GPT-Image-2 passes 90.4\% of images, whereas Qwen-Image and NextStep-1.1 pass only 12.7\% and 10.2\%. Qwen-Image-2.0 and GLM-Image succeed roughly half the time. Concept-cluster bootstrap comparisons resolve four performance tiers: the Qwen-Image-2.0/GLM-Image and Qwen-Image/NextStep-1.1 gaps are not statistically significant (Appendix~\ref{app:bootstrap}).

\noindent\textbf{Cognitive category predicts difficulty.} Figure~\ref{fig:category} reveals a systematic gradient: physical-spatiotemporal-quantitative concepts score highest (0.641), followed by emotion-sensation-experience (0.542); the three intermediate categories score 0.504--0.511, and interpersonal communication scores lowest (0.364). Even GPT-Image-2 drops from 0.91 on physical and emotion concepts to 0.83 on interpersonal communication; Qwen-Image-2.0 and GLM-Image reach only 0.25 and 0.30 on the latter. Spatial schemas and perceptible objects directly support physical concepts, and expression or posture provides cues for emotion. Communicative concepts such as \emph{excuse}, \emph{criticism}, and \emph{explanation} instead require speech acts and intent. Word-level scores follow this pattern: \emph{dream}, \emph{color}, and \emph{structure} reach 0.98, 0.97, and 0.96, whereas \emph{disagreement}, \emph{capital}, and \emph{excuse} score 0.11, 0.14, and 0.17. Visualization difficulty thus depends on cognitive structure, motivating category-balanced evaluation.

\noindent\textbf{Human validation and metric behavior.} On 90 images scored independently by three annotators, mean inter-annotator Spearman correlation is 0.622; automatic--human correlation is 0.362, with 91.1\% agreement within one rating bin (Table~\ref{tab:human}, Appendix~\ref{app:human-validation}). The rubric exposes semantic components that holistic image--text similarity misses: its correlation with image--concept cosine similarity is $\rho=-0.35$ overall and only 0.06--0.23 within models. These different behaviors motivate feature-level scoring of abstract meaning.

\noindent\textbf{Human imageability only partly predicts model success.} Across 150 concepts, human imageability has a weak positive association with mean adequacy ($\rho=0.215$, $p=0.008$). The association is stronger for physical concepts ($\rho=0.285$) and quality/mentality/belief ($\rho=0.237$), but nearly absent for emotion ($\rho=0.011$) and interpersonal communication ($\rho=0.030$). Thus ease of forming a mental image does not fully explain generation performance: the model must also select a depiction that preserves the target's semantic structure (Appendix~\ref{app:imageability}).

\noindent\textbf{Model comparisons are stable under rubric reweighting.} Equal-weight and human-calibrated scores correlate at $\rho=0.996$ across 2,700 images and 0.986 across 150 concepts. All six models retain the same ranking, with mean score changes at most 0.007 (Appendix~\ref{app:weighting}). The capability gaps therefore persist across these two weighting schemes, while calibration retains the relative importance assigned to individual semantic components.

\subsection{Representational Diversity: Prototype Reliance and Pseudo-Diversity}
\label{sec:diversity-results}

\begin{table}[t]
\centering
\small
\setlength{\tabcolsep}{4pt}
\begin{tabular}{lccccc}
\hline
Model & $D_{img}$ & $D_{text}$ & $H$ & Pseudo & Low \\
\hline
Qwen-Image & \textbf{0.297} & \textbf{0.280} & 1.15 & 10.7\% & 26 \\
NextStep-1.1 & 0.286 & 0.198 & 0.97 & \textbf{31.3\%} & 37 \\
GLM-Image & 0.274 & 0.262 & \textbf{1.28} & 14.7\% & 19 \\
Gemini-3.1-Flash & 0.246 & 0.209 & 1.01 & 20.0\% & 49 \\
Qwen-Image-2.0 & 0.228 & 0.194 & 0.92 & 12.0\% & 65 \\
GPT-Image-2 & 0.173 & 0.138 & 0.52 & 6.0\% & \textbf{112} \\
\hline
\end{tabular}
\caption{Diversity metrics by model (900 model$\times$concept groups in total; 150 per model). $D_{img}$/$D_{text}$ are mean pairwise within-group image/metaphor-embedding distances, $H$ is metaphor entropy, Pseudo is the pseudo-diversity rate, and Low is the number of groups falling in the low-image-distance $+$ low-metaphor-diversity quadrant.}
\label{tab:diversity}
\end{table}

\begin{figure*}[t]
\centering
\includegraphics[width=\linewidth]{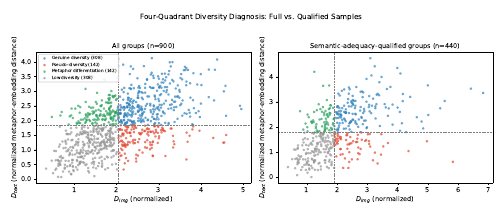}
\caption{Normalized image and metaphor-text distances for all 900 groups (left) and 440 adequacy-qualified groups (right). Dashed lines show the median splits used by the diagnostic.}
\label{fig:quadrants-full-qualified}
\end{figure*}

Figure~\ref{fig:quadrants-full-qualified} makes the distinction between surface and metaphor variation explicit. Among 900 model--concept groups, genuine diversity and low diversity each account for 34.2\%, while pseudo-diversity and metaphor differentiation each account for 15.8\%. Image and metaphor distances correlate at approximately 0.54: visual variation only partly tracks changes in expressed metaphor.

\begin{figure}[t]
\centering
\includegraphics[width=\linewidth]{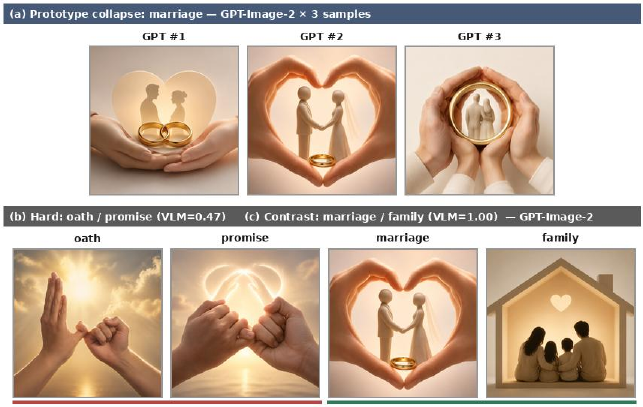}
\caption{Core failure modes. (a) Three GPT-Image-2 generations for \emph{marriage} repeat one prototype. (b) \emph{Oath/promise} shares visual motifs (pair-level VLM accuracy 0.47). (c) \emph{Marriage/family} has distinct visual carriers (1.00). Images are from GPT-Image-2; accuracies summarize all six models.}
\label{fig:casestudies}
\end{figure}

\noindent\textbf{High adequacy can coexist with prototype collapse.} GPT-Image-2 leads in semantic adequacy but has the lowest image distance, metaphor distance, and metaphor entropy (Table~\ref{tab:diversity}); 112 of its 150 groups fall in the low-diversity quadrant. It repeatedly maps \emph{marriage} to wedding rings and hearts and \emph{freedom} to a bird in flight (Figure~\ref{fig:casestudies}a). High semantic hit rates therefore leave a second capability unresolved: exploring multiple valid visual realizations of the same concept.

\noindent\textbf{Surface variation conceals repeated metaphors.} NextStep-1.1 combines near-highest image distance with the highest pseudo-diversity rate (31.3\%). Style and composition change while the underlying metaphor remains fixed. GLM-Image likewise shows high image distance but zero metaphor entropy for \emph{tradition} and \emph{good fortune}. Image-embedding distance alone overstates effective diversity in these cases.

\noindent\textbf{Adequacy filtering leaves pseudo-diversity intact.} In the 440 groups with at least two passing images, mean image distance falls from 0.251 to 0.226 and metaphor entropy from 0.97 to 0.76, yet pseudo-diversity barely changes (15.8\% to 15.5\%; Figure~\ref{fig:quadrants-full-qualified}). Semantic correctness therefore does not remove repeated metaphors. For cross-model comparison we use the full pool, since the filtered pool contains 136 GPT-Image-2 groups but only 15 NextStep-1.1 groups. Recurrent cases include \emph{tradition}, \emph{good fortune}, \emph{generation gap}, and \emph{celebration}, whose conventional symbols offer a plausible explanation for prototype reliance.

\subsection{Conceptual Distinctiveness: Fine-Grained Precision for Near-Synonyms}
\label{sec:distinctiveness-results}

\begin{table}[t]
\centering
\small
\setlength{\tabcolsep}{4pt}
\begin{tabular}{lcccc}
\hline
Model & Img LOO & MMD & Txt LOO & 2AFC \\
\hline
NextStep-1.1 & \textbf{0.883} & 0.440 & 0.850 & 0.728 \\
GPT-Image-2 & 0.861 & \textbf{0.454} & \textbf{0.878} & 0.821 \\
Qwen-Image-2.0 & 0.717 & 0.347 & 0.800 & 0.761 \\
GLM-Image & 0.650 & 0.288 & 0.733 & 0.839 \\
Gemini-3.1-Flash & 0.611 & 0.271 & 0.772 & 0.833 \\
Qwen-Image & 0.517 & 0.183 & 0.689 & \textbf{0.906} \\
\hline
Overall & 0.706 & 0.330 & 0.787 & 0.815 \\
\hline
\end{tabular}
\caption{Near-synonym distinctiveness results (30 pairs $\times$ 6 models; chance level 0.5). Img/Txt LOO denote leave-one-out nearest-centroid classification accuracy in image/text embedding space; VLM 2AFC is the primary perceptual-side metric.}
\label{tab:separability}
\end{table}

\noindent\textbf{Fine semantic boundaries remain difficult.} Across 180 model--pair results, mean VLM 2AFC accuracy is 0.815 and image-embedding leave-one-out accuracy is 0.706, both above chance (Table~\ref{tab:separability}). Pair-level variation is substantial. The difficult pairs \emph{economy/capital}, \emph{belief/faith}, and \emph{oath/promise} share visual motifs; \emph{oath/promise} reaches only 0.472 VLM accuracy. Oath-taking gestures, handshakes, and contract imagery fail to convey the distinction between a solemn public commitment and a general promise. Pairs with different visual carriers, such as \emph{space/distance} and \emph{conflict/violence}, are easier. This extends the challenge identified in knowledge-intensive T2I evaluation \citep{sun2026}: depicting a concept and preserving its boundary with a neighboring concept are distinct demands.

\noindent\textbf{Embedding separation and perceptual distinction diverge.} Qwen-Image has near-chance image-embedding accuracy (0.517) but the highest VLM accuracy (0.906); NextStep-1.1 shows the reverse pattern (0.883 and 0.728). Distributional separation and readable semantic cues therefore capture different properties. We use VLM 2AFC as the primary distinctiveness measure and embedding separation as a complementary diagnostic; text-embedding results broadly follow the image-side ranking.

\subsection{Case Studies}
\label{sec:cases}
Figure~\ref{fig:casestudies} connects the quantitative findings to generated content. Repeated wedding-ring imagery illustrates prototype collapse; overlapping gestures obscure \emph{oath/promise}, while wedding imagery versus household scenes distinguishes \emph{marriage/family} (pair-level accuracy 1.00). The central challenge is selecting a scene that expresses the distinguishing semantic components. Appendix~\ref{app:cases} expands these examples across repeated generations and models.

\subsection{What the Three Axes Reveal}
\label{sec:discussion}
The three axes expose different weaknesses in the same generation process. At the model--concept level, adequacy correlates negatively with image diversity ($\rho=-0.286$ over 900 groups); across the six model means, the correlation reaches $-0.886$. In contrast, adequacy and perceptual distinctiveness show little association across 180 model--pair results ($\rho=-0.044$, $p=0.55$; Appendix~\ref{app:statistics}). A model's success at expressing isolated concepts consequently does not determine whether it preserves their boundaries or explores alternative realizations.

This distinction has concrete implications for model improvement. For models with low adequacy, the first target is expressing the semantic relations identified by the rubric. For high-adequacy models with low metaphor diversity, improving surface style alone leaves prototype reliance unresolved; generation must explore additional valid situations and metaphors. For difficult near-synonyms, the target is a scene choice that makes the distinguishing component visible. ACVA-Bench provides separate measurements for these goals, turning a broad judgment of image quality into a diagnosis of where abstract meaning is lost.

\FloatBarrier
\section{Conclusion}
\label{sec:conclusion}
ACVA-Bench establishes a framework for evaluating the visual grounding of abstract meaning beyond explicit prompt following. Dictionary-grounded, human-reviewed rubrics make semantic adequacy measurable without prescribing a single depiction. Across six models and 2,700 images, we identify three recurring challenges: category-dependent semantic gaps, prototype reliance hidden by surface variation, and failures to preserve fine-grained concept boundaries. The three-axis evaluation exposes differences that image quality or a single aggregate score cannot capture, providing concrete targets for improving abstract-concept visualization.

\section*{Limitations}
ACVA-Bench uses 150 Chinese abstract concepts and three generations per model--concept pair; broader language coverage and larger sample sets would extend the evaluation. Automatic judgments are validated on 90 human-rated images. Additional annotators and independent judges would strengthen this validation. The current benchmark assesses static images under a standardized word-only prompt; richer prompting and temporal visual media provide natural directions for extension.

\section*{Ethical considerations}
Abstract concepts admit culturally diverse depictions. Our human-reviewed rubrics allow multiple visual realizations, and diversity analysis identifies reliance on recurring prototypes. Distribution of lexical resources, generated images, and human judgments should respect source licenses and provider terms and protect annotator privacy. These considerations guide interpretation and responsible use of the benchmark.

\bibliography{custom}

\appendix
\section{Reproducibility and Evaluation Implementation}
\label{app:implementation}

\paragraph{Generation protocol.}
For each of the 150 target concepts, all six models receive the same low-intervention template:
\begin{quote}
The target concept is ``\{word\}.'' Please generate an image that expresses this concept. Do not include any text in the image.
\end{quote}
Only the target word is substituted. No negative prompt, style modifier, aspect-ratio hint, or few-shot exemplar is supplied. Each model is called independently three times per concept using its default, non-seeded sampling configuration, yielding 450 images per model and 2,700 images overall. Seeds are not fixed because the objective is to characterize variation across independent calls, rather than reproduce a particular sample. The benchmark package contains the generated images, scores, and metadata for tracing individual outputs.

\paragraph{Semantic-adequacy judging.}
Qwen3.6-27B receives one image, the target concept, and its ordered visual-semantic question list (Section~\ref{sec:rubric}), with the following instruction:
\begin{quote}
You are shown an image intended to depict the concept ``\{concept\}.'' For each of the following visual-semantic questions, answer whether the image satisfies the condition, using exactly one of: 0 = not expressed, 0.5 = partially or indirectly expressed, 1 = fully expressed. Judge only what is visually observable; do not assume information that the image does not depict. Question list: \{questions\}.
\end{quote}
Per-question scores are aggregated using the human-calibrated weights (Sections~\ref{sec:auditing} and~\ref{sec:adequacy}).

\paragraph{Conceptual-distinctiveness judging.}
GLM-5V-Turbo receives one image and two candidate concepts $\{A,B\}$ in randomized left/right order. It selects the better-fitting label without knowing either the generating model or the target word used for generation. Randomization reduces position bias.

\paragraph{Compute.}
All six generators and the four evaluation/embedding models---Qwen3.6-27B, Qwen3-VL-Embedding-8B, Qwen3-Embedding-8B, and GLM-5V-Turbo---are accessed through hosted inference APIs. No local training, fine-tuning, or GPU inference is performed. CPU-based preprocessing and postprocessing, including clustering, quadrant statistics, and correlations, take under an hour for the 2,700-image corpus. Provider-side inference hardware is not controlled by the study. Reproduction uses the model versions and generation metadata distributed with the benchmark.

\paragraph{Judge-family effects.}
The semantic-adequacy judge and two generators, Qwen-Image and Qwen-Image-2.0, belong to the same model family. In Table~\ref{tab:adequacy}, these generators rank fifth and third, respectively, while GPT-Image-2 scores highest. This ranking does not show a simple same-family advantage, but it cannot rule out evaluator bias. Embedding-based separability measures a different construct and uses a Qwen-family embedding model, so it is not an independent validation of semantic-adequacy rankings. A controlled multi-judge study with an unrelated VLM remains future work.

\section{Category Definitions and an Illustrative Rubric}
\label{app:resources}

\paragraph{Canonical category names.}
Table~\ref{tab:canonical-categories} maps the six categories to the shorter labels used in the figures. The examples follow the benchmark's category assignments.

\begin{table*}[t]
\centering
\small
\begin{tabular}{p{0.39\textwidth}p{0.24\textwidth}p{0.27\textwidth}}
\toprule
Canonical category & Figure shorthand & Example concepts \\
\midrule
Physical, spatiotemporal, and quantitative & Physical / Spatiotemporal & color, space, distance, weather \\
Emotion, sensation, and experience & Emotion / Experience & happiness, shame, charm, nostalgia \\
Quality, mentality, and belief & Qualities \& Ideas & loyalty, freedom, belief, faith \\
Social-situation & Social Situations & marriage, family, conflict, violence \\
Interpersonal-communication & Interpersonal Communication & disagreement, excuse, criticism, explanation \\
Terminological, institutional, and epistemic & Institutional / Cognitive & economy, capital, system, structure \\
\bottomrule
\end{tabular}
\caption{Canonical category names and example concepts. Each category contains 25 concepts. The displayed column order is also used in Table~\ref{tab:category-full}.}
\label{tab:canonical-categories}
\end{table*}

\paragraph{Near-synonym pairs.}
The pairs \emph{belief/faith}, \emph{oath/promise}, \emph{marriage/family}, \emph{economy/capital}, \emph{space/distance}, and \emph{conflict/violence} illustrate the range discussed in Section~\ref{sec:distinctiveness-results}. They are part of the 30-pair set, constructed with five pairs per sampling stratum. Relatedness is checked using \emph{Tongyici Cilin} subclasses or near-synonym discrimination references. Table~\ref{tab:all-pairs} provides the full pair list and mean discrimination scores; English glosses identify the Chinese target concepts.

\paragraph{Worked example: loyalty.}
Table~\ref{tab:loyalty-rubric} illustrates how dictionary semantics, componential analysis, frame semantics, and image schemas can produce checkable questions. These five questions and their qualitative priorities are illustrative, not the full operational rubric or its calibrated numeric weights. Questions involving temporal relations or intentions must be answered from visible evidence rather than inferred backstory.

\begin{table*}[t]
\centering
\small
\begin{tabular}{p{0.79\textwidth}p{0.12\textwidth}}
\toprule
Illustrative visual-semantic question & Priority \\
\midrule
Does the image depict an ongoing, temporally extended relationship or bond, rather than a single isolated event? & high \\
Does one party show sustained support or allegiance toward another, rather than mere physical proximity or companionship? & highest \\
Is there a directional or asymmetric relation (e.g., one figure oriented toward, protecting, or standing by another) rather than a neutral symmetric pairing? & medium \\
Is attachment or bonding signified through a container/link schema (e.g., linked hands, rings, or an enclosing gesture) rather than incidental adjacency? & medium \\
Does the image avoid depicting betrayal, abandonment, or breach, which would contradict the target sense? & medium \\
\bottomrule
\end{tabular}
\caption{Illustrative questions for \emph{loyalty}. Priorities reproduce the schematic example; actual scores use the numeric weights established by human review (Section~\ref{sec:auditing}).}
\label{tab:loyalty-rubric}
\end{table*}

\begin{table*}[p]
\centering
\small
\begin{tabular}{rlrr}
\toprule
Rank & Concept pair (English glosses) & Mean VLM & Mean LOO \\
\midrule
1 & oath / promise & 0.472 & 0.528 \\
2 & system / structure & 0.556 & 0.750 \\
3 & custom / folklore & 0.583 & 0.694 \\
4 & perseverance / resolve & 0.694 & 0.556 \\
5 & belief / faith & 0.722 & 0.528 \\
6 & anger / hatred & 0.722 & 0.806 \\
7 & morality / ethics & 0.750 & 0.639 \\
8 & charm / temperament & 0.750 & 0.778 \\
9 & disagreement / contradiction & 0.750 & 0.694 \\
10 & economy / capital & 0.778 & 0.500 \\
11 & humiliation / shame & 0.806 & 0.694 \\
12 & youth / childhood & 0.806 & 0.778 \\
13 & institution / regime & 0.806 & 0.750 \\
14 & finance / subsidy & 0.833 & 0.806 \\
15 & weather / climate & 0.833 & 0.778 \\
16 & space / distance & 0.833 & 0.917 \\
17 & contest / activity & 0.833 & 0.639 \\
18 & dream / ideal & 0.861 & 0.528 \\
19 & output / quantity & 0.861 & 0.750 \\
20 & criticism / evaluation & 0.889 & 0.694 \\
21 & explanation / excuse & 0.889 & 0.694 \\
22 & goodwill / affection & 0.889 & 0.722 \\
23 & pressure / force & 0.911 & 0.667 \\
24 & conflict / violence & 0.917 & 0.806 \\
25 & etiquette / ritual & 0.917 & 0.750 \\
26 & quantitative / qualitative change & 0.917 & 0.806 \\
27 & nostalgia / sorrow & 0.944 & 0.806 \\
28 & happiness / fortune & 0.944 & 0.639 \\
29 & rumor / public opinion & 0.972 & 0.722 \\
30 & marriage / family & 1.000 & 0.778 \\
\bottomrule
\end{tabular}
\caption{All 30 near-synonym pairs, ordered by mean VLM 2AFC accuracy across six models. LOO denotes image-embedding leave-one-out nearest-centroid accuracy.}
\label{tab:all-pairs}
\end{table*}

\section{Extended Quantitative Results}
\label{app:quantitative}

\begin{figure}[t]
\centering
\includegraphics[width=\linewidth]{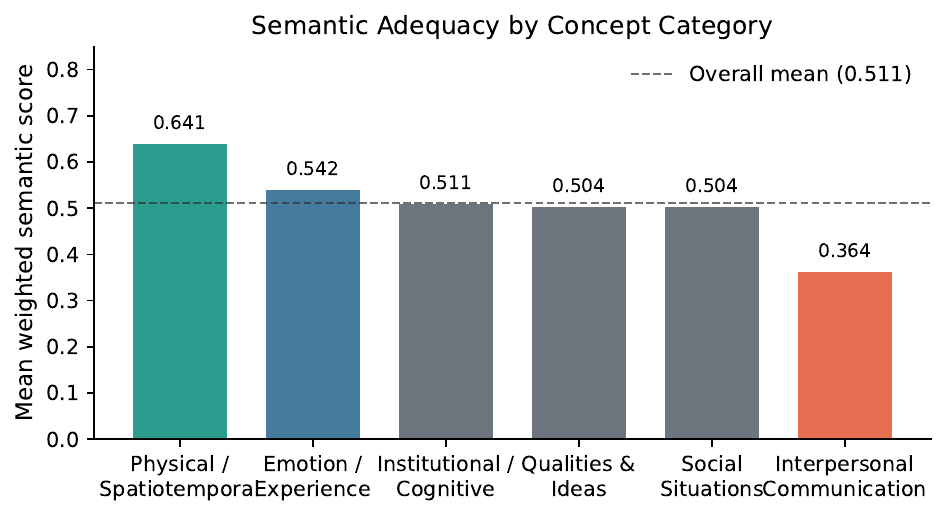}
\caption{Mean adequacy by cognitive category. The dashed line is the overall mean (0.511); canonical names are in Table~\ref{tab:canonical-categories}.}
\label{fig:category-bars}
\end{figure}

\paragraph{Category-level adequacy.}
Table~\ref{tab:category-full} reports model-by-category means to three decimal places. It expands Figure~\ref{fig:category} and the category analysis in Section~\ref{sec:experiments}. In canonical category order, the means are 0.641, 0.542, 0.504, 0.504, 0.364, and 0.511. Figure~\ref{fig:category-bars} displays the same means in descending order.

\begin{table*}[t]
\centering
\small
\begin{tabular}{lrrrrrr}
\toprule
Model & Phys. & Emot. & Qual. & Social & Interp. & Term. \\
\midrule
GPT-Image-2 & 0.907 & 0.913 & 0.872 & 0.891 & 0.830 & 0.834 \\
Gemini-3.1-Flash & 0.797 & 0.803 & 0.782 & 0.771 & 0.716 & 0.816 \\
Qwen-Image-2.0 & 0.712 & 0.629 & 0.580 & 0.562 & 0.254 & 0.597 \\
GLM-Image & 0.724 & 0.536 & 0.517 & 0.565 & 0.305 & 0.540 \\
Qwen-Image & 0.325 & 0.232 & 0.137 & 0.161 & 0.061 & 0.102 \\
NextStep-1.1 & 0.379 & 0.137 & 0.139 & 0.075 & 0.020 & 0.178 \\
\midrule
Category mean & 0.641 & 0.542 & 0.504 & 0.504 & 0.364 & 0.511 \\
\bottomrule
\end{tabular}
\caption{Semantic adequacy by model and category, rounded to three decimals; each cell averages 75 images. Columns follow Table~\ref{tab:canonical-categories}. Category means use the underlying unrounded values.}
\label{tab:category-full}
\end{table*}

\begin{figure}[t]
\centering
\includegraphics[width=\linewidth]{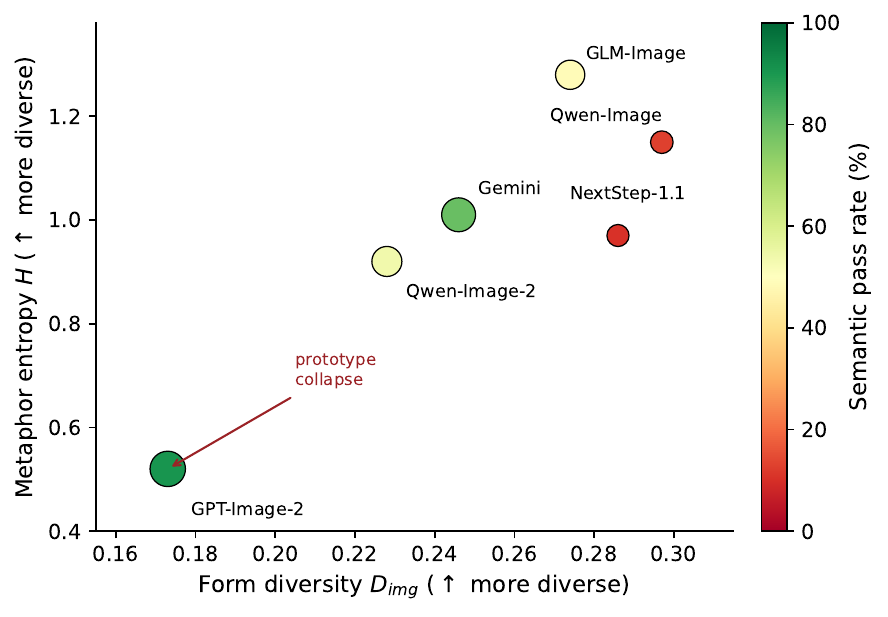}
\caption{Semantic pass rate versus diversity across six models. Axes show $D_{img}$ and metaphor entropy $H$; marker size and color encode the pass rate in Table~\ref{tab:adequacy}.}
\label{fig:diversity-tradeoff}
\end{figure}

\paragraph{Quality and diversity.}
Figure~\ref{fig:diversity-tradeoff} plots form diversity $D_{img}$ against metaphor entropy $H$, with marker size and color encoding semantic pass rate. GPT-Image-2 combines the highest pass rate with the lowest diversity on both axes. This illustrates the aggregate trade-off discussed in Section~\ref{sec:diversity-results}.

\paragraph{Full versus qualified groups.}
Figure~\ref{fig:quadrants-full-qualified} compares all 900 model--concept groups with the 440 groups containing at least two passing images. Pseudo-diversity changes from 15.8\% to 15.5\%; genuine and low diversity each change from 34.2\% to approximately 34.5\%. These similar aggregate proportions support the observation that adequacy filtering does not eliminate pseudo-diversity. They do not imply that the underlying groups or model composition are unchanged.

\begin{figure*}[p]
\centering
\includegraphics[width=0.82\linewidth]{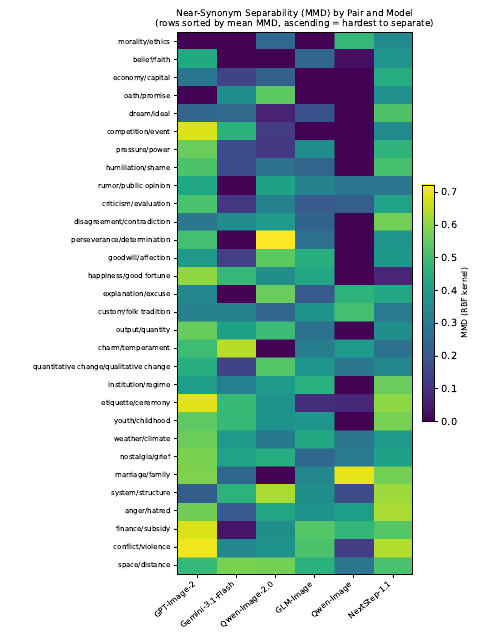}
\caption{MMD-based separability for 30 pairs and six models (RBF kernel), sorted by mean MMD. English labels gloss the Chinese concept pairs. This figure reports MMD, not perceptual 2AFC accuracy.}
\label{fig:pair-mmd}
\end{figure*}

\paragraph{Pair-level separability.}
Figure~\ref{fig:pair-mmd} expands the model-level MMD results to all 30 pairs. Rows are ordered by mean MMD across models. The five lowest-scoring pairs are \emph{morality/ethics}, \emph{belief/faith}, \emph{economy/capital}, \emph{oath/promise}, and \emph{dream/ideal}; \emph{conflict/violence} and \emph{space/distance} are among the five highest. MMD is an embedding-distribution measure; it should not be confused with the VLM 2AFC accuracy reported in Table~\ref{tab:all-pairs}.

\section{Supplementary Statistical Analyses}
\label{app:statistics}
The following analyses cover the 2,700 scored images, 900 model--concept diversity groups, and 180 model--pair distinctiveness results.

\subsection{Imageability and Semantic Adequacy}
\label{app:imageability}
\begin{figure}[t]
\centering
\includegraphics[width=\linewidth]{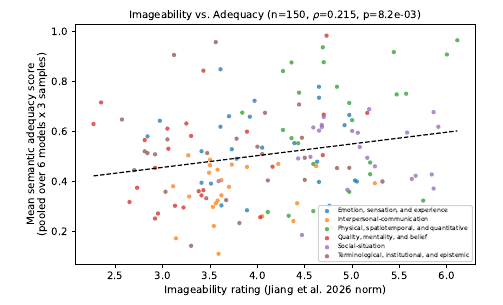}
\caption{Imageability versus mean semantic adequacy for 150 concepts, colored by category. The dashed line is an OLS fit; the reported association is Spearman $\rho=0.215$.}
\label{fig:imageability}
\end{figure}

Pooling six models and three generations gives 18 images per concept. Across 150 concepts, imageability from the source norms \citep{jiang2026} has a weak positive association with mean adequacy (Spearman $\rho=0.215$, $p=0.008$; Figure~\ref{fig:imageability}). Per-model correlations are $\rho=0.235$ ($p=0.004$) for GPT-Image-2, 0.219 ($p=0.007$) for GLM-Image, and 0.164 ($p=0.045$) for NextStep-1.1. The remaining estimates are 0.081 ($p=0.33$) for Gemini-3.1-Flash, 0.150 ($p=0.067$) for Qwen-Image-2.0, and 0.149 ($p=0.069$) for Qwen-Image. These are the reported, unadjusted significance values; no multiple-comparison correction is specified.

The relationship also varies by category: $\rho=0.285$ for physical/spatiotemporal/quantitative concepts and 0.237 for quality/mentality/belief, compared with 0.011 for emotion, 0.030 for interpersonal communication, 0.049 for social situations, and $-0.056$ for terminological/institutional/epistemic concepts. Human imageability therefore only weakly and unevenly predicts generation adequacy under this protocol.

\subsection{Relationships among Evaluation Axes}
\begin{figure}[t]
\centering
\includegraphics[width=\linewidth]{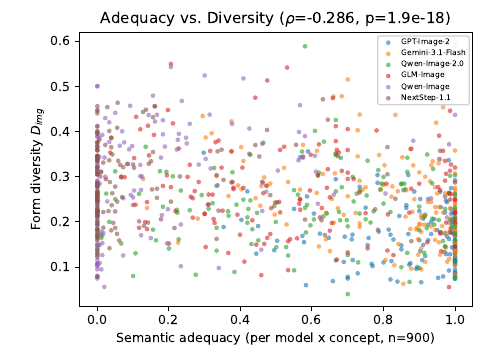}
\caption{Adequacy versus form diversity for 900 model--concept groups, colored by model ($\rho=-0.286$).}
\label{fig:adequacy-diversity}
\end{figure}

At the model--concept level ($n=900$), adequacy correlates negatively with form diversity ($\rho=-0.286$, $p=1.9\times10^{-18}$) and metaphor entropy ($\rho=-0.160$, $p=1.5\times10^{-6}$). Figure~\ref{fig:adequacy-diversity} visualizes the former. Across six model-level means, the adequacy--diversity correlation is stronger ($\rho=-0.886$), suggesting that between-model differences contribute substantially to the aggregate trade-off.

\begin{figure}[t]
\centering
\includegraphics[width=\linewidth]{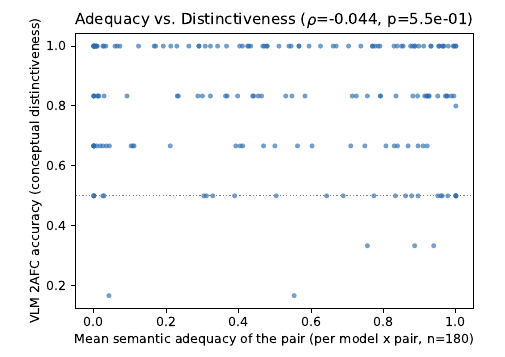}
\caption{Pair-level adequacy versus VLM 2AFC accuracy over 180 model--pair results ($\rho=-0.044$, $p=0.55$).}
\label{fig:adequacy-distinctiveness}
\end{figure}

At the model--pair level ($n=180$), mean adequacy and VLM 2AFC accuracy show little association ($\rho=-0.044$, $p=0.55$; Figure~\ref{fig:adequacy-distinctiveness}); the model-level correlation is similarly small ($\rho=0.029$). This supports measuring both axes, but a near-zero correlation does not establish statistical independence.


\subsection{Uncertainty in Model Comparisons}
\label{app:bootstrap}
\begin{figure}[t]
\centering
\includegraphics[width=\linewidth]{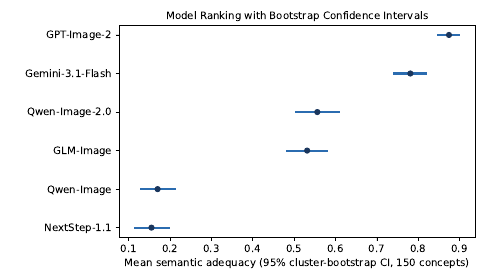}
\caption{Model adequacy with 95\% cluster-bootstrap intervals. Pairwise significance is assessed using difference tests in Appendix~\ref{app:bootstrap}.}
\label{fig:bootstrap}
\end{figure}

Table~\ref{tab:bootstrap} and Figure~\ref{fig:bootstrap} report 95\% cluster-bootstrap intervals for semantic adequacy, using 5,000 resamples of the 150 concepts. Resampling concepts keeps their three images together and accounts for within-concept dependence.

The reported adjacent-model comparisons identify three clear gaps: GPT-Image-2 versus Gemini-3.1-Flash, Gemini-3.1-Flash versus Qwen-Image-2.0, and GLM-Image versus Qwen-Image (reported bootstrap $p<0.0001$). Qwen-Image-2.0 versus GLM-Image has a difference interval of $[-0.045,0.094]$ ($p\approx0.49$), and Qwen-Image versus NextStep-1.1 has $[-0.042,0.073]$ ($p\approx0.61$). Thus the reported tests suggest four tiers rather than six statistically distinguishable ranks. Pairwise conclusions use these difference tests, rather than overlap of marginal confidence intervals alone. The precise tail-probability estimator for the reported $p$-values is not specified; intervals for diversity and distinctiveness are not reported.

\begin{table}[t]
\centering
\small
\begin{tabular}{lr}
\toprule
Model & Mean [95\% CI] \\
\midrule
GPT-Image-2 & 0.874 [0.848, 0.899] \\
Gemini-3.1-Flash & 0.781 [0.742, 0.818] \\
Qwen-Image-2.0 & 0.556 [0.504, 0.608] \\
GLM-Image & 0.531 [0.483, 0.579] \\
Qwen-Image & 0.170 [0.130, 0.213] \\
NextStep-1.1 & 0.155 [0.116, 0.196] \\
\bottomrule
\end{tabular}
\caption{Cluster-bootstrap intervals for mean adequacy (150 concept clusters; 5,000 resamples).}
\label{tab:bootstrap}
\end{table}

\subsection{Effect of Rubric Weighting}
\label{app:weighting}
Equal-weight and human-calibrated scores are highly correlated: Spearman $\rho=0.996$ over 2,700 images and 0.986 over 150 concepts. The six-model ranking is identical, with reported mean changes no larger than 0.007 (Table~\ref{tab:weighting}). Human calibration encodes the relative importance of semantic components, but these results show little effect on the aggregate comparisons; they do not demonstrate that weighting is necessary for the headline ranking.

\begin{table*}[t]
\centering
\small
\begin{tabular}{lrrr}
\toprule
Model & Unweighted & Weighted & Reported $\Delta$ \\
\midrule
GPT-Image-2 & 0.871 & 0.874 & +0.004 \\
Gemini-3.1-Flash & 0.776 & 0.781 & +0.005 \\
Qwen-Image-2.0 & 0.562 & 0.556 & -0.006 \\
GLM-Image & 0.535 & 0.531 & -0.004 \\
Qwen-Image & 0.170 & 0.170 & -0.000 \\
NextStep-1.1 & 0.162 & 0.155 & -0.007 \\
\bottomrule
\end{tabular}
\caption{Equal-weight versus human-calibrated mean adequacy. Reported deltas can differ from subtraction of the displayed rounded means.}
\label{tab:weighting}
\end{table*}

\section{Human Validation Protocol and Metric Interpretation}
\label{app:human-validation}

\begin{table*}[t]
\centering
\small
\begin{tabular}{lcccc}
\hline
Comparison & $n$ & Spearman $\rho$ & Exact & $\pm$1 \\
\hline
Inter-annotator (mean) & 90 & 0.622 & 0.696 & 0.970 \\
Auto vs.\ pooled human & 270 & 0.362 & 0.463 & 0.911 \\
\hline
\end{tabular}
\caption{Human validation of the weighted semantic score on 90 images. Auto = Qwen3.6-27B rubric scoring; inter-annotator Fleiss $\kappa=0.515$; Exact = exact agreement; $\pm$1 = agreement within one rating bin.}
\label{tab:human}
\end{table*}

\paragraph{Protocol.}
Three annotators independently evaluate the 90-image validation set using the same rubric and three-point question-level criterion as the automatic judge (Appendix~\ref{app:implementation}), followed by aggregation with the calibrated weights. Images and concept words are presented without the generating model's identity, automatic score, or other annotators' ratings. Mean pairwise inter-annotator Spearman correlation is 0.622 and Fleiss' $\kappa$ is 0.515 (Table~\ref{tab:human}), indicating incomplete agreement and interpretive variation even among humans.

\paragraph{Metric validity.}
The automatic--human correlation is $\rho=0.362$, exact agreement is 46.3\%, and agreement within one rating bin is 91.1\%. These results support a reproducible diagnostic signal, but do not establish a human-equivalent metric or guarantee preservation of model rankings. High adjacent-bin agreement should be interpreted alongside the exact-agreement rate and the three-level scoring scale. The automatic scores are slightly higher on average; a mean offset alone does not show that errors are uniform across concepts or models.

The weak or negative association with generic image--text similarity (Section~\ref{sec:experiments}) indicates that the measures behave differently, rather than independently proving rubric validity. Expanding beyond 90 images and three annotators, evaluating subgroup errors, and comparing unrelated judge models would strengthen the evidence for absolute-score interpretation.

\section{Extended Qualitative Examples}
\label{app:cases}

The main-text Figure~\ref{fig:casestudies}a shows a representative case of prototype collapse: for \emph{marriage}, GPT-Image-2's three samples converge on the same wedding-ring-and-heart schema, rarely exploring valid alternatives such as shared daily life; the same pattern recurs for other concepts (e.g., \emph{freedom}, rendered as a bird in flight). These depictions are not incorrect -- scores are often high -- but the model compresses the concept's semantic space into a single prototype, so repeated sampling brings no new perspective.

Figure~\ref{fig:casestudies}b--c contrasts near-synonym confusion with clear separation, using GPT-Image-2 images; the pair-level accuracies aggregate all six models. For the pair with the lowest VLM accuracy, \emph{oath/promise} (b), the two concepts share the same motifs -- an oath's solemnity and ceremonial setting are among the distinguishing semantics models fail to visualize (Section~\ref{sec:distinctiveness-results}) -- so accuracy falls to 0.47, below chance. By contrast, \emph{marriage/family} (c) reaches 1.00 accuracy, since wedding symbolism and family scenes are visually distinct. This connects near-synonym failures to whether the model translates distinguishing semes into a perceptible scene choice.

\subsection{Repeated Generations and Cross-Model Comparisons}
\begin{figure*}[t]
\centering
\includegraphics[width=\linewidth]{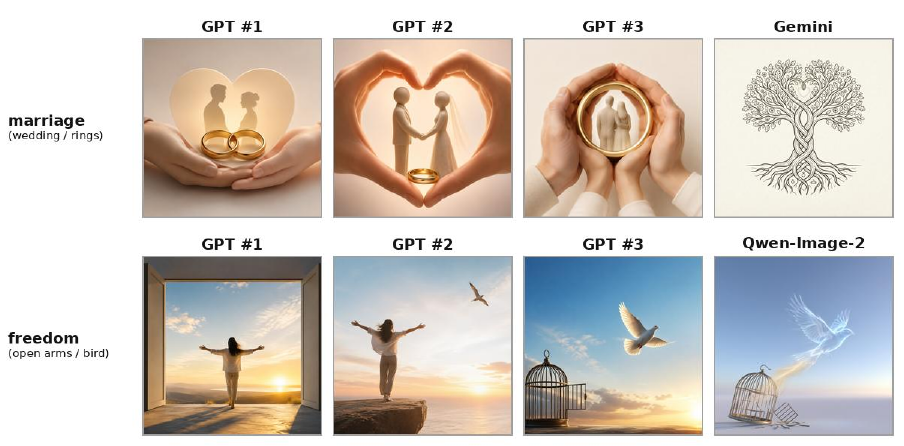}
\caption{Three independent GPT-Image-2 generations for \emph{marriage} and \emph{freedom}, with one contrasting model per row. Similar schemas across repeated samples illustrate prototype reliance.}
\label{fig:prototype-extended}
\end{figure*}

Figure~\ref{fig:prototype-extended} shows all three GPT-Image-2 generations for both \emph{marriage} and \emph{freedom}, together with one contrasting model per row. Wedding rings and hearts recur for marriage; open arms or released birds recur for freedom. These repeated schemas illustrate the low diversity of this model, while the alternative model outputs show that other depictions are possible.

\begin{figure*}[p]
\centering
\includegraphics[width=0.82\linewidth]{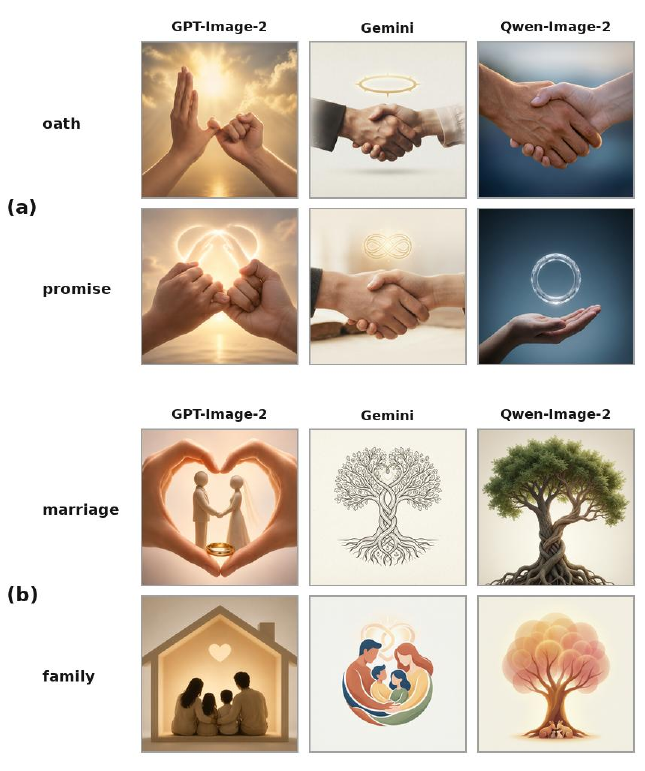}
\caption{Cross-model comparison for (a)~\emph{oath/promise} and (b)~\emph{marriage/family}. Three models illustrate recurring motif overlap in the harder pair and more distinct visual realizations in the easier pair.}
\label{fig:near-synonym-extended}
\end{figure*}

Figure~\ref{fig:near-synonym-extended} compares three models on \emph{oath/promise} and \emph{marriage/family}. The first pair shares motifs such as clasped hands, rings, or luminous gestures. The second contrasts romantic or wedding imagery with family or household scenes. These examples extend the single-model case study, without implying that every image or concept pair follows the same pattern.

\section{Supplementary Result Visualizations}
\label{app:plots}
The following plots provide complementary visualizations of the results in Tables~\ref{tab:adequacy} and~\ref{tab:separability}, with the category heatmap now shown in the main text (Figure~\ref{fig:category}). The corresponding results and interpretation are reported in Section~\ref{sec:experiments}.

\begin{figure*}[p]

\centering
\includegraphics[width=0.88\linewidth]{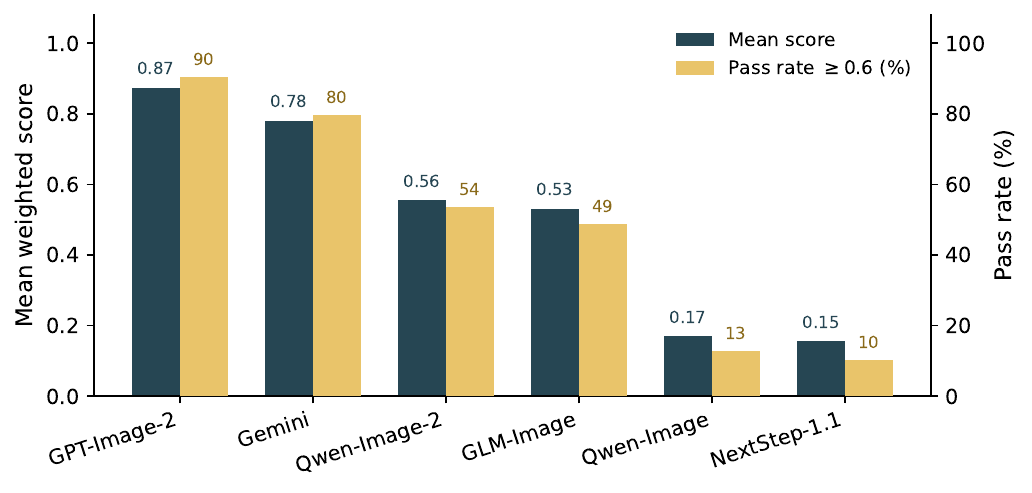}
\caption{Mean semantic adequacy and pass rate ($\geq$0.6 weighted score) by model.}
\label{fig:model_adequacy}

\centering
\includegraphics[width=0.88\linewidth]{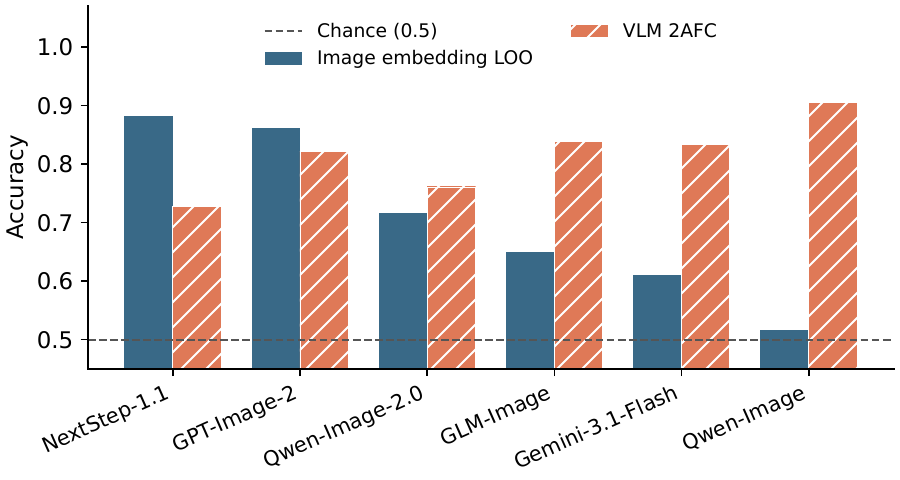}
\caption{Near-synonym distinctiveness: image-embedding leave-one-out accuracy versus VLM 2AFC accuracy. The dashed line is the chance level (0.5); the two rankings are not consistent with each other.}
\label{fig:separability}

\end{figure*}

\end{document}